\documentclass[11pt]{article}

\usepackage[final]{acl}
\usepackage{float}
\usepackage{times}
\usepackage{latexsym}
\usepackage{booktabs}
\usepackage[T1]{fontenc}

\usepackage[utf8]{inputenc}

\usepackage{microtype}

\usepackage{inconsolata}
\usepackage{graphicx}
\usepackage{amsmath}
\usepackage{amssymb}
\usepackage{tabularx}

\title{UCSC NLP at SemEval-2026 Task 10: Boundary-Aware Span Extraction and RoBERTa Classification for Conspiracy Detection}

\author{
  Dom Marhoefer \\
  UC Santa Cruz \\
  \texttt{dmarhoef@ucsc.edu} \And
  Milos Suvakovic \\
  UC Santa Cruz \\
  \texttt{msuvakov@ucsc.edu} \And
  Glenn Grant-Richards \\
  UC Santa Cruz \\
  \texttt{ggranri@ucsc.edu} \AND
  Aidan Pinero \\
  UC Santa Cruz \\
  \texttt{apinero@ucsc.edu} \And
  Ryan King \\
  UC Santa Cruz \\
  \texttt{rytking@ucsc.edu}
}

\begin{document}
\maketitle

\begin{abstract}
We present our systems for SemEval-2026 Task 10 (PsyCoMark), addressing conspiracy marker extraction (Subtask 1) and document-level conspiracy detection (Subtask 2).
For marker extraction, we formulate the task as multi-label span classification over enumerated candidate spans, using IoU$\ge$0.95 positive labeling, hard-negative sampling, and containment-based non-maximum suppression (NMS) with boundary-aware span representations.
Document classification is modeled independently using a sequence classifier with label smoothing and a stratified train--validation split. Analysis shows that entity-like roles (Actor, Victim) are detected robustly, while abstract roles (Action, Effect, Evidence) remain sensitive to boundary criteria.
On the official test set, our systems rank 7th in Subtask 1 (0.2251 macro F1) and 12th in Subtask 2 (0.7694 weighted F1).

\end{abstract}

\section{Introduction}

Conspiracy narratives are often described in terms of recurring narrative roles: a perceived \textit{Actor} performs an \textit{Action} against a \textit{Victim}, justified by \textit{Evidence} and producing harmful \textit{Effects}.
SemEval-2026 Task~10 (PsyCoMark) operationalizes this perspective through two subtasks:
extracting conspiratorial roles from text and classifying whether a document expresses a conspiracy narrative \citep{samory-etal-2026-semeval}.

\begin{itemize}
\item \textbf{Subtask~1: Conspiracy-marker extraction.}
Given a document, systems predict labeled spans corresponding to five semantic roles
(\textit{Actor}, \textit{Action}, \textit{Effect}, \textit{Evidence}, \textit{Victim}).
The output is a set of markers of the form \texttt{\{type, start, end\}} per document.

\item \textbf{Subtask~2: Document-level conspiracy classification.}
Each document is assigned one of three labels (\textit{Yes}, \textit{No}, \textit{Can't tell}).
\end{itemize}

A central feature of the PsyCoMark subtasks is that the previously defined role structures occur in both conspiratorial and non-conspiratorial discourse (critical narrative).
Documents labeled as non-conspiratorial may contain complete \textit{Actor}--\textit{Action}--\textit{Victim} structures without conspiratorial intent, while conspiratorial documents differ primarily in framing and epistemic stance. As a result, marker extraction and conspiracy detection are intrinsically related but not equivalent: success requires the precise boundary identification of abstract semantic roles while also modeling document-level stance.

For Subtask~1, we use multi-label span classification over enumerated candidate spans with IoU$\ge$0.95 labeling, hard-negative mining, and containment-based NMS; for Subtask~2, we use a document classifier with label smoothing and a stratified train--validation split.
On the SemEval evaluation server, our best submissions ranked 7th in Subtask~1 (macro F1 0.2251) and 12th in Subtask~2 (weighted F1 0.7694). Both models achieved competitive performance relative to other shared-task submissions.

Analysis indicates that boundary sensitivity remains a primary failure mode for abstract roles (\textit{Action}, \textit{Effect}, \textit{Evidence}), particularly under stricter token-level overlap criteria.
Entity-like roles (\textit{Actor}, \textit{Victim}) were comparatively robust, suggesting that semantic abstraction and variable span length dominate marker difficulty rather than topical content.

\section{Background and Task Setup}

\subsection{PsyCoMark Task and Data}

\paragraph{Dataset characteristics.}
The PsyCoMark corpus contains over 4{,}100 English Reddit submission statements drawn from more than 190 subreddits \citep{samory-etal-2026-semeval}. The document-level conspiracy labels are moderately imbalanced, with 35.3\% Yes, 46.6\% No, and 18.1\% Can't tell instances. Annotations are dense: the majority of documents contain at least one psycholinguistic marker, and many contain multiple role types.

A \textbf{non-conspiratorial} example contains multiple markers without endorsing a conspiracy:
\begin{quote}
``[Germany]$_{\textit{Actor}}$ has [upset]$_{\textit{Action}}$ [other]$_{\textit{Victim}}$ EU member states by securing a [disproportionately]$_{\textit{Effect}}$ large share of the bloc's common pool of vaccines, according to [a]$_{\textit{Evidence}}$ report\ldots''
\end{quote}

In contrast, a \textbf{conspiratorial} document expresses conspiratorial framing:
\begin{quote}
``So [they]$_{\textit{Actor}}$ want us to believe it was a [suicide]$_{\textit{Effect}}$, when [It's]$_{\textit{Evidence}}$ so blatantly obvious it's not\ldots just because [Jeffrey]$_{\textit{Victim}}$ is dead doesn't mean it ends here\ldots he was [under suicide watch]$_{\textit{Action}}$\ldots''
\end{quote}

This overlap of role structure across labels makes marker extraction and conspiracy classification intrinsically coupled yet non-identical tasks.

\subsection{Related Work and Modeling Motivation}

\paragraph{Psycholinguistic structure.}
Conspiracy thinking is associated with pattern seeking, perceived threat, and intentional plot framing \citep{wood2012dead,vanprooijen2017history}. Linguistically, this often manifests as oppositional discourse and in-group vs.\ out-group framing \citep{korencic2024what}, which computationally distinguishes conspiratorial from critical narratives. These findings suggest that conspiratorial narratives exhibit recurrent semantic role structure rather than purely topical content, motivating approaches that explicitly model roles such as Actor, Victim, and Evidence.

\paragraph{Computational approaches.}
Prior work has incorporated psycholinguistic features such as LIWC-derived lexical categories \citep{tausczik2010liwc} and broader psycholinguistic profiles to identify conspiracy propagators \citep{giachanou2023conspiracy}, or fine-tuned neural models for document-level conspiracy classification \citep{Liu_2025}. Hybrid systems combining emotional or moral framing features have also been explored \citep{george2024a}. While such features provide indirect cues about epistemic stance, they do not explicitly represent narrative role relations.

To analyze these relations, span-based classification has been widely adopted in information extraction tasks such as coreference resolution and semantic role labeling, where enumerating candidate spans enables modeling of overlapping and variable-length structures \citep{lee-etal-2017-end}.

\section{System Overview}

We address PsyCoMark with two independent RoBERTa-large \citep{liu2019roberta} models, one for each subtask. Rather than incorporating external psycholinguistic features, our model
learns role representations directly from the span-level supervision
provided by PsyCoMark annotations. Subtask~1 is formulated as multi-label classification over enumerated candidate spans, with post-processing to resolve overlapping and nested predictions.
Subtask~2 is formulated as standard sequence classification over the full document text. We train the two models separately and perform no parameter sharing or cross-task feature transfer; outputs are combined only to match the required submission format.
Experiments were conducted on a single NVIDIA A100 GPU and a single RTX-4070 GPU using PyTorch 2.10 and Hugging Face Transformers 5.1.

\subsection{Subtask 1: Conspiracy Marker Extraction}

\subsubsection{Architecture}

A key challenge in PsyCoMark span extraction is precise boundary localization for semantically abstract, multi-word roles whose lengths vary substantially across contexts. We formulate conspiracy marker extraction as span classification rather than token-level tagging.
Given a tokenized document, we enumerate all candidate spans up to a maximum length of $L=32$ tokens (chosen for efficiency).
Each candidate span $(i,j)$ is independently classified into one or more semantic roles.

For each span $(i,j)$, we construct a contextual representation
$v_{span}$ by concatenating six components:
\[
v_{span} = [h_i ; h_j ; \bar{h}_{i:j} ; w_{emb} ; h_{i-1} ; h_{j+1}]
\]

where:
\begin{itemize}
\item $h_i, h_j \in \mathbb{R}^H$: contextualized embeddings of the span start and end tokens.
\item $\bar{h}_{i:j} \in \mathbb{R}^H$: mean-pooled contextualized embedding over tokens within the span.
\item $w_{emb} \in \mathbb{R}^H$: learned embedding encoding span width (length). The embedding table size is $33 \times H$.
\item $h_{i-1}, h_{j+1} \in \mathbb{R}^H$: contextualized embeddings of the immediate left and right context tokens (zero-padded at boundaries).
\end{itemize}

With RoBERTa-large hidden size $H=1024$, the concatenated representation has dimension $6H = 6144$. The span vector is passed through a two-layer MLP, producing logits for the five conspiracy roles:
\[
6H \xrightarrow{\text{Linear}} H \xrightarrow{\text{ReLU}} \xrightarrow{\text{Dropout}(0.1)} \xrightarrow{\text{Linear}} 5
\]

\subsubsection{Training}

\paragraph{Loss Function.}
We train the span classifier using binary cross-entropy with logits (BCE) for multi-label prediction.
To address severe class imbalance (most candidate spans are negative), we apply per-role positive weighting:
\[
\text{pos\_weight}_r = \frac{N_{neg,r}}{N_{pos,r}}
\]
where $N_{neg,r}$ and $N_{pos,r}$ are the counts of negative and positive spans for role $r$, computed from the first 200 training batches to approximate class frequency without a full pass over the training set. We clip weights at 20.0 for stability.

\paragraph{Labeling Strategy.}
To enforce precise boundary learning, a candidate span is labeled positive for a role only if its Intersection-over-Union (IoU) with a gold span is $\ge 0.95$.
This high-overlap criterion encourages exact span boundaries rather than approximate matches.

\paragraph{Sampling.}
Enumerating all spans yields an overwhelming number of negative examples.
We therefore use a three-tier sampling strategy to construct each training batch (up to 160 spans per document):

\begin{enumerate}
\item \textbf{Positives:} All gold-aligned spans.
\item \textbf{Hard negatives:} Spans with partial overlap (IoU $\in [0.50, 0.75)$) with gold spans.
\item \textbf{Random negatives:} Additional non-overlapping spans sampled uniformly.
\end{enumerate}

Hard negatives introduce near-boundary confusions that improve span boundary discrimination.

\subsubsection{Decoding}

Span predictions are converted into final markers through three post-processing steps:

\begin{enumerate}
\item \textbf{Thresholding:}
We apply per-role probability thresholds to sigmoid scores.
For the submitted run, thresholds were tuned on the validation set via per-role grid search to maximize Macro F1 under \emph{token}-level IoU $\ge 0.5$ (matching the official evaluation).

\item \textbf{Containment-based NMS:}
Overlapping spans are pruned using non-maximum suppression (NMS).
A lower-scoring span $B$ is suppressed by a higher-scoring span $A$ if
their containment ratio $\frac{|A \cap B|}{\min(|A|,|B|)}$
exceeds $\text{contain\_thr}$, or if $\text{IoU}(A,B) \ge \text{iou\_thr}$.
This removes redundant nested or highly overlapping predictions.
We use role-specific thresholds (contain\_thr 0.65--0.75, iou\_thr 0.35--0.45) tuned on the validation set, with more permissive values for low-recall roles (Action, Effect, Evidence) to improve recall.

\item \textbf{Span merging:}
After NMS, adjacent spans of the same role separated by a small character gap ($\le 3$) are merged to form contiguous markers.
\end{enumerate}

\subsubsection{Experimental Details}

\paragraph{Hyperparameters.}
We fine-tune RoBERTa-large with AdamW using a learning rate of $2\times10^{-5}$,
weight decay 0.01, linear warmup over 10\% of training steps, and cosine decay thereafter.
The maximum sequence length is 512 tokens and the maximum span length is 32.
Training uses batch size 2 for up to 14 epochs with early stopping
(patience 5) based on validation decoded micro F1 (character IoU $\ge$ 0.3). We use random seed 42.

\subsubsection{Results}

On the validation set, our best span model achieved a decoded micro F1 of
\textbf{0.6501} after thresholding and containment-based NMS, where a predicted span is
counted correct if it matches a gold span with character-level IoU $\ge 0.3$.
Per-role validation performance is shown in Table~\ref{tab:subtask1_results}.

\begin{table}[H]
\centering
\begin{tabular}{lccc}
\toprule
Role & Precision & Recall & F1 \\
\midrule
Actor & 0.745 & 0.798 & 0.771 \\
Action & 0.656 & 0.527 & 0.584 \\
Effect & 0.603 & 0.491 & 0.541 \\
Evidence & 0.652 & 0.504 & 0.569 \\
Victim & 0.744 & 0.644 & 0.691 \\
\midrule
\textbf{Micro Avg} & \textbf{0.692} & \textbf{0.613} & \textbf{0.650} \\
\bottomrule
\end{tabular}
\caption{Subtask 1 validation results (decoded spans; character IoU $\ge 0.3$).}
\label{tab:subtask1_results}
\end{table}
On the official test set, the shared task reports token-based overlap F1
with IoU $\ge 0.5$. Under this evaluation, our submitted run achieved 0.2251 macro F1 and 0.2408
micro F1.

The difference between validation and official scores reflects both the stricter token-level IoU threshold and the greater sensitivity of macro F1 to low-frequency roles. Validation used IoU $\ge 0.3$ during development to stabilize span boundary learning, while submission thresholds were tuned under IoU $\ge 0.5$ to match the shared-task metric.

\begin{table}[H]
\centering
\begin{tabular}{lccc}
\toprule
Role & Precision & Recall & F1 \\
\midrule
Actor & 0.252 & 0.684 & 0.368 \\
Action & 0.106 & 0.331 & 0.160 \\
Effect & 0.098 & 0.256 & 0.142 \\
Evidence & 0.128 & 0.293 & 0.178 \\
Victim & 0.191 & 0.508 & 0.277 \\
\midrule
\textbf{Macro F1} & \multicolumn{3}{c}{0.225} \\
\bottomrule
\end{tabular}
\caption{Subtask 1 official test results (token IoU $\ge 0.5$).}
\end{table}

\subsection{Subtask 2: Conspiracy Detection}

\subsubsection{Architecture}

We formulate Subtask~2 as document-level sequence classification.
Each input document is encoded with RoBERTa-large, and the contextual
representation of the \texttt{[CLS]} token is used as a fixed-length
document embedding.

This representation is passed through a classification head consisting
of a linear projection with Tanh activation and dropout, followed by a
final linear layer producing logits for the three labels:
\textit{Yes}, \textit{No}, and \textit{Can't tell}. The RoBERTa encoder and classification head are jointly fine-tuned.

\subsubsection{Training}

\paragraph{Loss Function.}
We train the document classifier using cross-entropy loss over the three
labels (\textit{Yes}, \textit{No}, \textit{Can't tell}).
To reduce overconfidence and improve generalization under class imbalance,
we apply label smoothing with $\alpha = 0.05$.

\paragraph{Data Split.}
We use a stratified 90/10 train--validation split to preserve class
distribution.

\paragraph{Optimization.}
The model is fine-tuned with AdamW using a linear learning-rate schedule
with warmup.
Training proceeds for up to 8 epochs with early stopping (patience 3)
based on validation Weighted F1.

\subsubsection{Experimental Details}

\paragraph{Hyperparameters.}
We fine-tune RoBERTa-large with AdamW using a learning rate of $2\times10^{-5}$,
weight decay 0.01, and linear warmup over 10\% of training steps.
The maximum sequence length is 512 tokens and label smoothing is 0.05.
Training uses batch size 4 for up to 8 epochs with early stopping
(patience 3).

\subsubsection{Results}

On the validation split, our best classifier achieved Weighted F1 0.6739
and Accuracy 0.6744.
On the official test set, our submitted run achieved Weighted F1 0.7694
and Accuracy 0.7730. The official evaluation script excludes instances labeled ``Can't tell,'' resulting in metrics computed over 608 matched samples. These scores are therefore not directly comparable to our three-class validation results. The higher test performance likely reflects both this label exclusion and differences in split difficulty.

\section{Discussion}

\subsection{What Worked}

\paragraph{Hard Negative Mining.}
Hard negatives (spans with 50--75\% overlap with gold spans) proved critical
for boundary learning. Without these near-miss examples, the model frequently
predicted overly broad spans that contained the correct marker but included
extraneous context. Hard negatives encouraged sharper span localization.

\paragraph{Subtask Independence.}
The document-level classifier in Subtask~2 receives only the raw input
text and does not use predicted spans or role features from
Subtask~1. Outputs from the two subtasks are combined only at submission time to
match the shared-task output format. No information is exchanged between
models during training or inference. This design isolates span extraction and document classification performance and avoids cross-task error propagation.

\paragraph{Evaluation Alignment.}
Performance was highly sensitive to how closely development
metrics matched the official evaluation protocol.
For Subtask~1, early model selection used a relaxed character-level
IoU criterion to stabilize boundary learning, whereas the shared-task
evaluation uses token-level overlap with a stricter IoU threshold.
This mismatch explains the validation--test gap and highlights the
importance of calibrating span learning directly to the target metric.
For Subtask~2, the official scoring excluded ``Can't tell'' instances,
changing the effective evaluation distribution relative to three-class
validation. Aligning validation procedures with the exact scoring
setup proved as important as architectural choices.

\subsection{Limitations}
\paragraph{Error Analysis.}
Under the strict token $\text{IoU} \ge 0.5$ evaluation metric, our model achieves 0.65 micro F1 on the validation split. The decrease to 0.225 macro F1 on the official test set indicates substantial distribution shift in the unseen data and the outsized impact of low-frequency abstract roles.

Qualitative error analysis reveals two dominant boundary failure modes. False positives frequently manifest as boundary drift, where the model captures extraneous syntactic modifiers (e.g., predicting \texttt{[April 5 statement]} instead of the precise gold \textit{Evidence} annotation \texttt{[statement]}). Conversely, false negatives stem from substantial under-segmentation of compositional phrases (e.g., predicting just \texttt{["nuclear demolition".]} instead of the full \textit{Action} clause \texttt{[spilled the beans on the Mossad operation involving "nuclear demolition".]}).

\paragraph{Abstract Role Boundaries.}
Roles such as \textit{Action}, \textit{Effect}, and \textit{Evidence} exhibit
greater semantic abstraction and variable span length than entity-like roles
(\textit{Actor}, \textit{Victim}). This led to lower recall and boundary
inconsistency, particularly for longer or compositional spans.
The large drop from relaxed validation IoU (0.3) to official IoU (0.5)
indicates substantial boundary sensitivity; future work should evaluate
with stricter metrics during training to better align with shared-task
evaluation.

\paragraph{Span Enumeration Constraints.}
Limiting candidate spans to length $L=32$ improved computational efficiency
but may exclude valid long markers, especially for multi-clause \textit{Effect}
or \textit{Evidence} spans. The fixed maximum span length therefore imposes a
recall ceiling.

\subsection{Future Work}

\paragraph{Joint Modeling of Subtasks.}
While independence avoided error propagation, well-implemented joint modeling could allow
document classification to leverage predicted role structure. For example,
aggregated \textit{Actor} and \textit{Victim} spans could provide explicit
narrative signals for Subtask~2.

\paragraph{Structured Span Decoding.}
Current decoding relies on thresholding and NMS heuristics. Structured
prediction approaches (e.g., span graphs or conditional decoding) could better
enforce role consistency and reduce overlapping errors.

\paragraph{Role-Aware Encoding.}
Incorporating role-conditioned representations or span-type embeddings during
encoding may improve discrimination of abstract roles and reduce boundary
ambiguity.

\paragraph{Psycholinguistic and Lexical Features.}
While our current architecture relies exclusively on learned contextual embeddings, incorporating psycholinguistic markers, sentiment analysis, and topic modeling could help better disentangle conspiratorial framing from critical narrative.

\section{Conclusion}

\paragraph{Official evaluation summary.}
On the SemEval evaluation server, our system achieved 0.2251 macro F1 for span extraction and 0.7694 weighted F1 for document classification.
For Subtask~1, span classification with hard-negative training and
containment-based decoding achieved 0.6501 decoded micro F1 on validation (character IoU $\ge$ 0.3, internal development metric) and 0.2251 macro F1 on the official test.
For Subtask~2, a standard RoBERTa-large classifier achieved 0.6739 weighted F1 on validation and 0.7694 on test
without explicit role features. These results ranked 7th in Subtask 1 and 12th in Subtask 2 among participating systems, indicating that document-level conspiracy
signals can be learned directly from raw text.
Future work includes joint modeling of span roles and document classification
to better capture narrative structure.

\bibliography{custom}

\end{document}